\documentclass[11pt,a4paper]{article}
\usepackage[hyperref]{acl2020}
\usepackage{times}
\usepackage{latexsym}
\usepackage[draft]{todonotes}   

\usepackage{graphicx}
\usepackage{makecell}
\graphicspath{./images/}

\usepackage{microtype}

\aclfinalcopy 
\def\aclpaperid{1958} 

\makeatletter
\def\blfootnote{\gdef\@thefnmark{}\@footnotetext}
\makeatother

\title{From Arguments to Key Points: Towards Automatic\\Argument Summarization }
\author{ Roy Bar-Haim \qquad Lilach Eden \qquad Roni Friedman \qquad Yoav Kantor\\ 
 \textbf{Dan Lahav \qquad Noam Slonim$^{*}$}\\
				 IBM Research\\
				\texttt{\{roybar,lilache,roni.friedman-melamed,yoavka,noams\}@il.ibm.com}\\
				\texttt{dan.lahav@ibm.com}}

\date{}

\begin{document}
\maketitle
\begin{abstract}
Generating a concise summary from a large collection of arguments on a given topic is an intriguing yet understudied problem. We propose to represent such summaries as a small set of talking points, termed \emph{key points}, each scored according to its salience. We show, by analyzing a large dataset of crowd-contributed arguments, that a small number of key points per topic is typically sufficient for covering the vast majority of the arguments. Furthermore, we found that a domain expert can often predict these key points in advance. We study the task of argument-to-key point mapping, and introduce a novel large-scale dataset for this task. We report empirical results for an extensive set of experiments with this dataset, showing promising performance. 
\end{abstract}
\section{Introduction}
\blfootnote{
    \hspace{-0.2cm}
		$^{*}$All authors equally contributed to this work.
}
\begin{table*}[t!]\small
    \begin{center}
\begin{tabular}{p{13cm} r}
\multicolumn{1}{c}{\textbf{Homeschooling should be banned}}& \#Args\\
\hline
\hline
 \emph{Pro}	&	\\
 \hline
Mainstream schools are essential to develop social skills.	&	61	\\
Parents are not qualified as teachers.	&	20	\\
Homeschools cannot be regulated/standardized.	&	15	\\
Mainstream schools are of higher educational quality.	&	9	\\
\hline
\emph{Con}	&		\\
\hline
Parents should be permitted to choose the education of their children.	&	28	\\
Homeschooling is often the best option for catering for the needs of exceptional/religious/ill/disabled students.	&	25	\\
Homeschools can be personalized to the child's pace/needs.	&	21	\\
Mainstream schools have a lot of violence/bullying.	&	21	\\
The home is a good learning environment.	&	13	\\
Parents will have more ability to pay-attention/educate their child.	&	7	\\
\hline
\hline
    \end{tabular}
    \caption{A sample key point-based summary, extracted from our ArgKP dataset.}
    \label{tab:summary}

    \end{center}
\end{table*}
Governments, businesses and individuals, all need to make decisions on a daily basis: \emph{``Should cannabis be legalized?''},\emph{``Should we develop this product?''}, \emph{``Should I become a vegetarian?''}. When making an important decision, the process typically comprises several steps: first, we gather as much information as we can about the pros and cons of the proposal under consideration. We may then summarize the collected information as a short list of the main arguments for each side. Lastly, we aim to weigh the pro and con arguments against each other to make the final decision.

Where can we find relevant arguments for a given topic? In recent years, significant progress was made in the field of \emph{argument mining}, automatic identification and extraction of argumentative structures in text \citep{lawrence-reed-cl-2019}. Specifically, several works focused on topic-related argument mining from the Web or other massive corpora \citep{Levy:ARGMINING2017,Levy:COLING2018,Wachsmuth:ARGMINING2017,Stab:NAACL2018,Stab:EMNLP2018,eindor2019corpus}. Policy makers in governments or businesses may also conduct surveys to collect from large audiences arguments supporting or contesting some proposal. 

Each of the above methods may result in hundreds or thousands of arguments per topic, making it impossible for the decision maker to read and digest such large amounts of information. Several works aimed to alleviate this problem by clustering together related arguments, based on different notions of relatedness, such as \emph{similarity}  \citep{reimers-etal-2019-classification}, \emph{frames} \citep{ajjour-etal-2019-modeling}, and \emph{argument facets} \citep{misra-etal-2016-measuring}. These works, however, did not attempt to create a concise textual summary from the resulting clusters. 

In this work we propose to summarize the arguments supporting each side of the debate by mapping them to a short list of talking points, termed \emph{key points}. The salience of each key point can be represented by the number of its matching arguments. An example for such summary is shown in Table~\ref{tab:summary}. Key points may be viewed as high-level arguments. They should be general enough to match a significant portion of the arguments, yet informative enough to make a useful summary. 

The proposed method raises a fundamental question: can a small number of key points effectively summarize massive amount of arguments collected from a large population? In this work we give a positive answer to this question, based on extensive analysis over 28 controversial topics and 7,000 crowd-contributed pro and con arguments for these topics. Furthermore, we found that, given a controversial topic, a domain expert can compose a short, comprehensive list of key points even without looking at the arguments themselves.

Motivated by the above findings, we assume in this work that the key points for each topic are given, and focus on the task of automatically mapping arguments to these key points. This setting may be viewed as an intermediate step towards fully automatic argument summarization, but also as a valuable setting by itself: argument-to-key point mapping allows measuring the distribution of key points in a massive collection of arguments. It also allows interactive exploration of large argument collections, where key points serve as queries for retrieving matching arguments. In addition, it can be used for novelty detection - identifying unexpected arguments that do not match presupposed key points. 

We develop the \emph{ArgKP} dataset for the argument-to-keypoint mapping task, comprising about 24,000 \emph{(argument, key point)} pairs labeled as matching/non matching.\footnote{The dataset is available at \url{https://www.research.ibm.com/haifa/dept/vst/debating_data.shtml}} To the best of our knowledge, this is the first dataset for this task. As discussed in the next section in more detail, our dataset is also much larger and far more comprehensive than datasets developed for related tasks such as mapping posts or comments in online debates to \emph{reasons} or \emph{arguments} \citep{hasan-ng-2014-taking,boltuzic-snajder-2014-back}. 

We report empirical results for an extensive set of supervised and unsupervised configurations, achieving promising results.

The main contributions of this work are:
\begin{enumerate}
\item We demonstrate, through extensive data annotation and analysis over a variety of topics, the feasibility and effectiveness of summarizing a large set of arguments collected from a large audience by mapping them to a small set of key points. 
\item We develop the first large-scale dataset for the task of argument-to-key point mapping.
\item We perform empirical evaluation and analysis of a variety of classification methods for the above task.
\end{enumerate}
\section{Related Work}
\subsection{Argument Mining}
The starting point for the current work is a collection of pro and con arguments for a given topic. As previously mentioned, these arguments may be collected from a large audience by conducting a survey, or mined automatically from text. 

Some of the previous work on argument mining focused on specific domains such as legal documents \citep{Moens:2007,Wyner:2010}, student essays \citep{stab-gurevych-2017-parsing, persing-ng-2016-end}, and user  comments  on  proposed  regulations \cite{park-cardie-2014-identifying}. 

Mining arguments and argument components for a given topic (also known as \emph{context}) has been a prominent line of research in argument mining. \citet{levy-etal-2014-context} introduced the task of context-dependent claim detection in a collection of Wikipedia articles, and \citet{rinott-etal-2015-show} did the same for context-dependent evidence detection. More recently, several works focused on topic-related argument mining from the Web or other massive corpora \citep{Levy:ARGMINING2017,Levy:COLING2018,Wachsmuth:ARGMINING2017,Stab:NAACL2018,Stab:EMNLP2018,eindor2019corpus}.

Stance classification of extracted arguments can be performed as a separate step \cite{Barhaim:2017} or jointly with argument detection, as a three-way classification (pro argument/con argument/none), as done by \citet{Stab:EMNLP2018}.
\subsection{Argument Clustering and Summarization}
\label{ssec:argrel}
Several works have focused on identifying pairs of similar arguments, or clustering similar arguments together.  \citet{ajjour-etal-2019-modeling} addressed the task of splitting a set of arguments
into a set of non-overlapping \emph{frames} such as \emph{Economics}, \emph{Environment} and \emph{Politics}.  \citet{reimers-etal-2019-classification} classified argument pairs as similar/dissimilar.  \citet{misra-etal-2016-measuring} aimed to detect argument pairs that are assumed to share the same \emph{argument facet}, which is similar to our notion of \emph{key points}. However, they did not attempt to explicitly identify or generate these facets, which remained implicit, but rather focused on detecting similarity between argument pairs. In contrast to these works, we directly map arguments to key points.

\citet{egan-etal-2016-summarising} proposed to summarize argumentative discussions through the extraction of salient ``points'', where each point is a verb and its syntactic arguments. Applying their unsupervised method to online political debates showed significant improvement over a baseline extractive summarizer, according to human evaluation. While the current work also aims to summarize argumentative content via concise points, our goal is not to extract these points but to accurately map arguments to given points. Our main challenge is to identify the various ways in which the meaning of a point is conveyed in different arguments. The method employed by \citeauthor{egan-etal-2016-summarising} only matches arguments with the same \emph{signature} - the same verb, subject and object dependency nodes, hence its ability to capture such variability is limited.

The line of work that seems most similar to ours is of \citet{hasan-ng-2014-taking}, \citet{boltuzic-snajder-2014-back} and \citet{naderi-2016}. \citeauthor{hasan-ng-2014-taking} classified posts and individual sentences from online debates into a closed set of \emph{reasons}, composed manually for each topic. \citeauthor{boltuzic-snajder-2014-back} mapped comments from one debating website (\emph{ProCon.org}) to arguments taken from another debating website (\emph{iDebate.org}). \citet{naderi-2016} addressed a similar task: she used part of the \citeauthor{boltuzic-snajder-2014-back} corpus as training data for an SVM classifier, which was then tested on sentences and paragraphs from same-sex marriage debates in the Canadian Parliament, annotated with the same set of arguments.

Our work differs from these works in several respects. First, we deal with crowd-contributed arguments, taken from the dataset of \citet{gretz2019largescale} while these works dealt with posts or comments in debate forums, and parliamentary debates. Second, the dataset developed in this work is far more extensive, covering 28 topics and over 6,500 arguments\footnote{As detailed in the next section, a few hundreds of arguments out of the initial 7,000 were filtered in the process of constructing the dataset.}, as compared to 2-4 topics in the datasets of \citeauthor{boltuzic-snajder-2014-back} and \citeauthor{hasan-ng-2014-taking}, respectively. This allows us to perform a comprehensive analysis on the feasibility and effectiveness of argument-to-key point mapping over a variety of topics, which has not been possible with previous datasets. Lastly, while \citeauthor{hasan-ng-2014-taking} only perform within-topic classification, where the classifier is trained and tested on the same topic, we address the far more challenging task of cross-topic classification. \citeauthor{boltuzic-snajder-2014-back} experimented with both within-topic and cross-topic classification, however they used a limited amount of data for training and testing: two topics, with less than 200 comments per topic.

Finally, we point out the similarity between the argument/key point relation and the text/hypothesis relation in \emph{textual entailment}, also known as \emph{natural language inference (NLI)} \citep{DBLP:series/synthesis/2013Dagan}. Indeed, \citet{boltuzic-snajder-2014-back} used textual entailment as part of their experiments, following the earlier work of \citet{cabrio:ac13}, who used textual entailment to detect support/attack relations between arguments. 

\section{Data}
\subsection{Arguments and Key Points}
\label{arguments_and_kps}

As a source of arguments for this work we have used the publicly available IBM-Rank-30k dataset \cite{gretz2019largescale}. This dataset contains around 30K crowd-sourced arguments, annotated for polarity and point-wise quality. The arguments were collected
with strict length limitations, accompanied by extensive quality control measures. Out of the 71 controversial topics in this dataset, we selected the subset of 28 topics for which a corresponding motion exists in the \emph{Debatabase} repository of the \emph{iDebate} website\footnote{\url{https://idebate.org/debatabase}}. This requirement guaranteed that the selected topics were of high general interest. 

We filtered arguments of low quality (below 0.5) and unclear polarity (below 0.6), to ensure sufficient argument quality in the downstream analysis. We randomly sampled 250 arguments per topic from the set of arguments that passed these filters (7,000 arguments in total for the 28 topics).


Debatabase lists several pro and con points per motion, where each point is typically 1-2 paragraphs long. The headline of each point is a concise sentence that summarizes the point. Initially, we intended to use these point headlines as our key points. However, we found them to be unsuitable for our purpose, due to a large variance in their level of specificity, and their low coverage of the crowd's arguments, as observed in our preliminary analysis.   

To overcome this issue, we let a domain expert who is a professional debater write the key points from scratch. The expert debater received the list of topics and was asked to generate a maximum of 7 key points for each side of the topic, without being exposed to the list of arguments per topic. 
The maximal number of key points was set according to the typical number of pro and con points in Debatabase motions. 

The process employed by the expert debater to produce the key points comprised several steps: 
\begin{enumerate}
	\item Given a debate topic, generate a list of possible key points in a constrained time frame of 10 minuets per side.
	\item Unify related key points that can be expressed as a single key point.
	\item Out of the created key points, select a maximum of 7 per side that are estimated to be the most immediate ones, hence the most likely to be chosen by crowd workers. 
\end{enumerate}

The process was completed within two working days. A total of 378 key points were generated, an average of 6.75 per side per topic.  
%
%
%
%
\subsection{Mapping Arguments to Key Points}
\subsubsection{Annotation Process}

Using the Figure Eight crowd labeling platform\footnote{\url{http://figure-eight.com}}, we created gold labels for associating the arguments selected as described in Section~\ref{arguments_and_kps} with key points. For each argument, given in the context of its debatable topic, annotators were presented with the key points created for this topic in the relevant stance. 
They were guided to mark all of the key points this argument can be associated with, and if none are relevant, to select the 'None' option. Each argument was labeled by $8$ annotators.

\textbf{Quality Measures:} to ensure the quality of the collected data, the following measures were taken - \begin{enumerate}
    \item Test questions. Annotators were asked to determine the stance of each argument towards the topic. Similarly to \citet{toledo-etal-2019-automatic}, this question functioned as a hidden text question\footnote{Unlike \citeauthor{toledo-etal-2019-automatic}, the results were analyzed after the task was completed, and the annotators were not aware of their success/failure.}. All judgments of annotators failing in more than $10\%$ of the stance questions were discarded. 
    \item Annotator-$\kappa$ score. This score, measuring inter annotator agreement, as defined by \citet{toledo-etal-2019-automatic}, was calculated for each annotator, and all judgments of annotators with annotator-$\kappa < 0.3$ were ignored. This score averages all pair-wise Cohen's Kappa \citep{landis+koch77} for a given annotator, for any annotator sharing at least $50$ judgments with at least $5$ other annotators.
    \item Selected group of trusted annotators. As in \citet{gretz2019largescale}, the task was only available to a group of annotators which had performed well in previous tasks by our team. 
\end{enumerate}
As described above, the annotation of each key point with respect to a given argument was performed independently, and each annotator could select multiple key points to be associated with each given argument. For the purpose of calculating inter-annotator agreement, we considered \emph{(argument, key point)} pairs, annotated with a binary label denoting whether the argument was matched to the key point. 
Fleiss' Kappa for this task was $0.44$ \citep{fleiss:71}, and Cohen's Kappa was $0.5$ (averaging Annotator-$\kappa$ scores). These scores correspond to ``moderate agreement'' and are comparable to agreement levels previously reported for other annotation tasks in computational argumentation \citep{boltuzic-snajder-2014-back,eindor2019corpus}. As for the stance selection question, $98\%$ of the judgments were correct, indicating overall high annotation quality. 

\textbf{Data Cleansing}: In addition to the above measures, the following annotations were removed from the data: (i) Annotations in which the answer to the stance selection question was wrong; (ii) Annotations in which key point choice was illegal -  the 'None' option and one of the key points were both selected. However, the rate of these errors, for each of the annotators, was rather low ($<10\%$ and $<5\%$, respectively).

Arguments left with less than $7$ valid judgments after applying the above quality measures and data cleansing were removed from the dataset. $6,568$ labeled arguments remain in the dataset. 
%
\subsubsection{Annotation Results}
\label{annotations_results}
\begin{table*}[t]\small
\begin{center}
\begin{tabular}{|p{3.5cm}|p{5.5cm}|p{5.5cm}|} 
\hline
\multicolumn{1}{|c|}{\textbf{Topic}} &
\multicolumn{1}{|c|}{\textbf{Argument}} & \multicolumn{1}{|c|}{\textbf{Associated Key Point(s)}}  \\ 
\hline
\hline
We should end mandatory retirement. &Forcing members of a profession to retire at a certain age creates an experience drain.&A mandatory retirement age decreases institutional knowledge. \\ 
\hline
We should ban the use of child actors. &Child actors are fine to use as long as there is a responsible adult watching them.&Child performers should not be banned as long as there is supervision/regulation. \\ 
\hline
We should close Guantanamo Bay detention camp.&Guantanamo can provide security for accused terrorists who would be hurt in the general prison population.&The Guantanamo bay detention camp is better for prisoners than the alternatives. \\ 
\hline
Assisted suicide should be a criminal offence.&People have a basic right to bodily autonomy, deciding whether or not to die with minimal suffering and dignity is integral to that right.&
    People should have the freedom to choose to end their life.\newline
    Assisted suicide gives dignity to the person that wants to commit it.
\\ 
\hline
We should ban human cloning.&The world is already overpopulated, cloning humans will only contribute to this problem.&No key point \\ 
\hline
\end{tabular}
\caption{Examples for key point association to arguments.}
\label{tab:label_type_examples}
\end{center}
\end{table*}
Next, we consolidate the individual annotations as follows. We say that an argument $a$ is mapped to a key point $k$ if at least 60\% of the annotators mapped $a$ to $k$. Recall that an argument can be mapped to more than one key point. Similarly, we say that $a$ has \emph{no key point} if at least 60\% of the annotators mapped $a$ to None (which is equivalent to not selecting any key point for the argument). Otherwise, we say that $a$ is \emph{ambiguous}, i.e., the annotations were indecisive. Table~\ref{tab:label_type_examples} shows examples for arguments and their matching key points in our dataset. 
 
The distribution of the arguments in the dataset over the above categories is shown in Table~\ref{tab:num_kp_per_arg}. Remarkably, our key points, composed independently of the arguments, were able to cover 72.5\% of them, with 5\% of the arguments mapped to more than one key point. 

We further investigated the differences between arguments in each category, by comparing their average quality score (taken from the IBM-Rank-30k dataset), number of tokens and number of sentences. The results are shown as additional columns in Table~~\ref{tab:num_kp_per_arg}. Interestingly, arguments that have no key point tend to be shorter and have lower quality score, comparing to arguments mapped to a single key point; arguments mapped to more than one key point are the longest and have the highest quality. 

\begin{table*}[]\small
\centering
\begin{tabular}{|l|r|r|r|r|}
\hline
 & \% Arguments &Quality & \# Tokens & \# Sentences \\ \hline \hline
No key point & 4.7\% & 0.75 & 16.35 & 1.09 \\ \hline
Ambiguous & 22.8\% & 0.80 & 18.97 & 1.15 \\ \hline
Single key point & 67.5\% & 0.84 & 18.54 & 1.15 \\ \hline
Multiple key points & 5.0\% & 0.91 & 23.66 & 1.33 \\ \hline
\end{tabular}
\caption{Argument statistics by key point matches.}
\label{tab:num_kp_per_arg}
\end{table*}
%
%
%

\begin{figure}[t!]
  \includegraphics[scale=0.7]{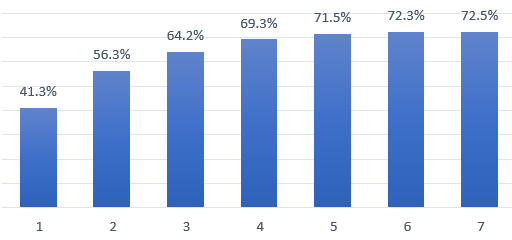}
  \caption{Argument coverage per number of key points.}
  \label{fig:avg_kp_coverage}
\end{figure}

Figure~\ref{fig:avg_kp_coverage} examines the impact of the number of key points on argument coverage. For each topic and stance, we order the key points according to the number of their matched arguments, and add them incrementally. The results indicate that arguments are not trivially mapped to only one or two key points, but a combination of several key points is required to achieve high coverage. The marginal contribution decays for the sixth and seventh key points, suggesting that seven key points indeed suffice for this task. 

22.8\% of the arguments are \emph{ambiguous}. Annotations for these arguments are split over several possible key points, none reaching the 60\% threshold. For instance, the argument \emph{``homeschooling enables parents with fringe views to push their agenda on their children without allowing exposure to alternative viewpoints.''}, had two key points with annotator votes higher than 40\%, but below 60\%:
\begin{enumerate}
\item \emph{Homeschools cannot be regulated / standardized.}
\item \emph{Parents are not qualified as teachers.}
\end{enumerate}
Such cases suggest that many arguments are somewhat covered by the key points, but if the judgment is not clear-cut, the different intuitions of the annotators may result in no label receiving the required majority.     
%
%
%
\subsection{Final Dataset Generation}
The \emph{ArgKP} dataset includes \emph{(argument, key point)} pairs with binary labels indicating whether the argument is matched to the key point. The dataset was created from the labeled data as follows. We define the \emph{label score} of a pair as the fraction of annotations that classified the pair as matching . Pairs with label score $\geq 0.6$ were labeled as positive (matching). Pairs with label score $\leq 0.15$ were labeled as negative (non-matching). Pairs with label score in between these thresholds were removed. 

We further cleansed our data by discarding key points having less than three matching arguments. This led to the removal of 135 out of the 378 key points and 14,679 out of 38,772 pairs obtained from the previous step.

The final dataset has 24,093 labeled \emph{(argument, key point)} pairs, of which 4,998 pairs (20.7\%) are positive. It has 6,515 arguments (232.67 per topic), and 243 key points (8.67 key points per topic). For each pair, the dataset also specifies the topic and the stance of the argument towards the topic.

We assessed the quality of the resulting dataset by having an expert annotator\footnote{A professional debater who was not involved in the development of the dataset.} mapping 100 randomly sampled arguments to key points, and comparing the annotations to the gold labels for all the corresponding pairs in the dataset.  We obtained a remarkably high Cohen's Kappa of 0.82 (``almost perfect agreement''), validating the high quality of the dataset.
\section{Experiments}
\subsection{Experimental Setup}
We perform the task of matching arguments to key points in two steps. In the \emph{Match Scoring} step (Section~\ref{sec:match_scoring}), we generate a score for each argument and key point. Then, in the \emph{Match Classification} step (Section~\ref{sec:match_classification}), we use these scores to classify the pairs as matching or non-matching.  

We perform 4-fold cross-validation over the ArgKP dataset. Each fold comprises 7 test topics, 17 train topics and 4 development topics.

\subsubsection{Match Scoring}
\label{sec:match_scoring}
We experimented with both unsupervised and supervised methods for computing a match score for a given \emph{(argument, key point)} pair. We also explored transfer learning from the related task of natural language inference (NLI).
\paragraph{Unsupervised Methods}
\begin{itemize}
\item \textbf{Tf-Idf.} In order to assess the role of lexical overlap in the matching task, we represent each argument and key point as tf-idf weighted word vectors and use their cosine similarity as the match score.

\item \textbf{Word Embedding.} We examined averaged word embeddings using GloVe \citep{pennington2014glove} and BERT \citep{devlin-etal-2019-bert}. GloVe is a context independent model that computes a single embedding for each word. BERT is a contextualized embedding model that takes the entire sentence into account. We also experimented with other embedding methods that under-performed BERT and thus their results are not reported here: Universal Sentence Encoder \citep{cer2018universal} and InferSent \citep{conneau2017supervised}. Again, we use cosine similarity to compute the match score.
%
\end{itemize}

\paragraph{Supervised Methods.} 
We fine tuned the BERT-base-uncased and BERT-large-uncased models \citep{devlin-etal-2019-bert} to predict matches between argument and key point pairs. We added a linear fully connected layer of size 1 followed by a sigmoid layer to the special [CLS] token in the BERT model, and trained it for three epochs with a learning rate of 2e-5 and a binary cross entropy loss.

\paragraph{NLI Transfer Learning.} We also experimented with transfer learning from NLI to our task of argument-to-key point match classification. This was motivated by the similarity between these tasks (as discussed in Section~\ref{ssec:argrel}), as well as the availability of large-scale NLI labeled datasets. We considered the Stanford (SNLI) and the Multi-Genre (MNLI) datasets \citep{bowman-etal-2015-large,williams-etal-2018-broad}, each comprising hundreds of thousands of labeled premise-hypothesis pairs. Pairs labeled as \textsc{Entailment} were considered positive instances, while the rest of the pairs, labeled as \textsc{Neutral} or \textsc{Contradiction} were considered negative. We trained BERT-base and BERT-large models on each of these datasets, following the procedure described above. 
\subsubsection{Match Classification} \label{Selection Policy}
\label{sec:match_classification}
In the match classification step we select the matching key points for each argument, based on their respective matching scores.  
The classification can be done locally, treating each pair individually, or globally, by examining all possible key points for each argument. We compared the following policies for selecting matching key points for a given argument.

\paragraph{Threshold.} For each fold, we find the threshold on the match score that maximizes the F1 score for the positive (matching) class. Pairs whose score exceeds the learned threshold are considered matched. 

\paragraph{Best Match (BM).} Using a threshold is not optimal for our data, where most arguments have at most one matched key point. A natural solution is to select the best matching key point. For each argument, we consider all key points for the same topic and stance as candidates and predict only the candidate with the highest match score as matched to the argument and the rest as unmatched. Note that this is the only fully unsupervised selection policy, as it does not require labeled data for learning a threshold.

\paragraph{BM+Threshold.} The \emph{BM} policy always assigns exactly one key point for each argument, while 27.5\% of the arguments in our data are not matched to any key point. To address this, we combine the two former policies. The top matching key point is considered a match only if its match score exceeds the learned threshold.

\paragraph{Dual Threshold.} In order to account for arguments with more than one matching key point, two thresholds are learned. If two key points exceed the lower threshold and at least one of them exceeds the upper threshold, both will be matched. Otherwise, it works the same as the \emph{BM+Threshold} policy using only the lower threshold. This allows for zero to two matches per argument.

Thresholds are learned from the development set for supervised match scoring methods, and from both train and development set for unsupervised match scoring methods.
\subsection{Results}
\subsubsection{Match Scoring Methods}
\begin{table*}[t]\small
\begin{center}
\begin{tabular}{|c|c|cccc|}
\hline
\multicolumn{2}{|c|}{}                               & Acc & P & R &  F1    \\
\hline
                     & Majority Class      & 0.793  &       & 0.000     &       \\
                     & Random Predictions & 0.679 & 0.206 & 0.200   & 0.203 \\
\hline
Unsupervised Methods & Tf-Idf             & 0.512 & 0.246 & 0.644 & 0.352 \\
                     & Glove Embeddings   & 0.346 & 0.212 & 0.787 & 0.330  \\
                     & BERT Embeddings    & 0.660  & 0.319 & 0.550  & 0.403 \\
\hline
Supervised Methods   & BERT-base (ArgKP)  & 0.844 & 0.609 & 0.718 & 0.657 \\
                     & BERT-large (ArgKP) & \textbf{0.868} & \textbf{0.685} & \textbf{0.688} & \textbf{0.684 }
 \\
\hline
NLI Transfer Learning & BERT-base (SNLI)  & 0.777& 0.472 & 0.514& 0.485 \\
										 & BERT-base (MNLI)  & 0.772& 0.470 & 0.558& 0.505 \\
										 & BERT-large (SNLI) & 0.765& 0.456 & 0.533& 0.487 \\
										 & BERT-large (MNLI) & 0.792& 0.518 & 0.542& 0.526 \\ 
\hline
\end{tabular}
    \caption{Comparison of match scoring methods, using the \emph{Threshold} selection policy. P, R and F1 refer to the positive class. Acc is the accuracy.}
    \label{tab:Results}
\end{center}
\end{table*}
\begin{table*}[t]\small
\centering
\begin{tabular}{|c||c|c|c|c||c|c|c||c|c|c||c|}
\hline
& \multicolumn{4}{c||}{All} & \multicolumn{3}{c||}{Single}& \multicolumn{3}{c||}{Multiple}& No \\
& \multicolumn{4}{c||}{Arguments} & \multicolumn{3}{c||}{Key Point}& \multicolumn{3}{c||}{Key Points}& Key Points\\
\hline
              & Acc & P & R & F1             & P & R & F1               & P & R & F1         & Acc \\
\hline
Threshold    & .868 & .685 & .688 & .684 & .720  & .686 & .701 & .904 & \textbf{.690}  & \textbf{.782} & \textbf{.933} \\
Best Match & .876 & .696 & .711 & .703 & .836 & .747 & \textbf{.789} & .936 & .448 & .606 & .839 \\
BM+Threshold & \textbf{.890} & \textbf{.772} & .665 & .713 & \textbf{.856} & .699 & .769 & .941 & .421 & .580  & .915 \\
Dual Threshold & .887 & .721 & \textbf{.740}  & \textbf{.730}  & .784 & \textbf{.752} & .767 & \textbf{.945} & .656 & .773 & .908 \\
\hline
\end{tabular}
    \caption{Comparing key point selection policies, using BERT-large trained on the ArgKP dataset for match scoring.}
    \label{tab:Results-policy}
\end{table*}

Table~\ref{tab:Results} compares the various match scoring methods, all using the \emph{Threshold} key point selection policy. Results are obtained by micro-averaging over the argument-key point pairs in each fold, and averaging over the different folds. We consider Precision, Recall and F1 of the positive class, as well as the overall accuracy. We also list for reference the majority class baseline that always predicts ``no match'', and the random baseline, which randomly predicts the positive class according to its probability in the training data.

The unsupervised models fail to capture the relation between the argument and the key points. Tf-Idf and Glove perform the worst, showing that simple lexical similarity is insufficient for this task. BERT embedding does better but still reaches a relatively low F1 score of 0.4.

In contrast to the unsupervised models, supervised models are shown to perform well. BERT with fine tuning leads to a substantial improvement, reaching F1 score of 0.657 with the BERT-base model, and 0.684 with the BERT-large model. 

BERT Models trained on NLI data are considerably better than the unsupervised methods, with the best model reaching F1 of 0.526, yet their performance is still far below the supervised models trained on our ArgKP dataset. This may reflect both the similarities and the differences between NLI and the current task. We have also experimented with combining these two types of data in cascade: BERT was first trained on a large NLI dataset (SNLI, MNLI or their union), and was then fine-tuned on the smaller ArgKP data. However, it did not improve the supervised results. 	 
\paragraph{Error Analysis.}
By analyzing the top errors of the supervised classifier (BERT-large), we found several systematic patterns of errors. In most cases, non-matching arguments and key points received a high match score in one of the following cases: 
\begin{itemize}
\item They share some key phrases. For example: \emph{``It is unfair to only subsidize vocational education. Achieving a more advanced education is very expensive and it would also need to be subsidized.''} and \emph{``Subsidizing vocational education is expensive''.}    
\item They share a large portion of the sentence, but not the main point, for example: \emph{``Women should be able to fight if they are strong enough''} and \emph{``Women should be able to serve in combat if they choose to''.}
\item They are at least partially related, but labeled as non-matching due to a better fitting key point for the same argument. For example: \emph{``We should subsidize space exploration because it increases the knowledge of the universe we are in''} and \emph{``Space exploration improves science/technology''} can be considered matched, but were labeled as unmatched due to the key point \emph{``Space exploration unravels information about the universe''}. Using the \emph{Best Match} policy helps in these cases. 
\end{itemize}

For arguments and key points that were labeled as matched but received a low match score, the relation was in many cases implied or required some further knowledge, for examples: \emph{``Journalism is an essential part of democracy and freedom of expression and should not be subsidized by the state.''} and \emph{``government intervention has the risk of inserting bias/harming objectivity''.}
\subsubsection{Key Point Selection Policies}
\begin{figure*}[t]
   \centering
     \includegraphics[width=0.6\textwidth]{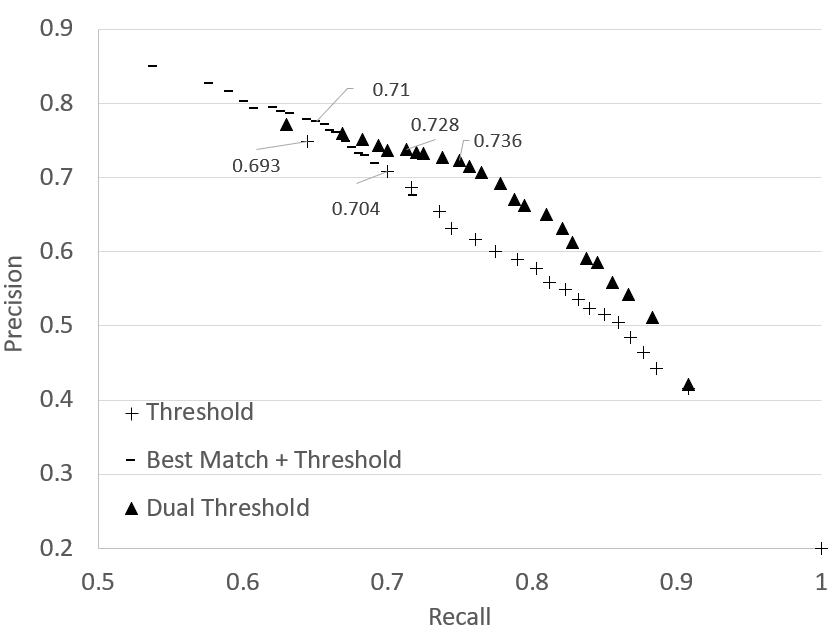}
     \caption{Precision/Recall trade-off for different key point selection policies. For each method, the highest F1 score, as well as the F1 score for the chosen threshold are specified. For the \emph{Best Match + Threshold} policy, these two  scores coincide.}
     \label{tab:prec-acc}
 \end{figure*}
 Table~\ref{tab:Results-policy} compares different key point selection policies, all using the best performing match scoring method: BERT-large fine-tuned on ArgKP. We report the results over the whole dataset (``all arguments''), as well as the subsets of arguments having none, single or multiple matching key points according to the labeled data. In case of no matches we present the accuracy, as recall and F1 scores are undefined. When considering all the arguments, the \emph{Dual Threshold} policy achieves the best F1 score of 0.73. The \emph{Threshold} method performs well for arguments with no matches or multiple matches. When there is exactly one match (the common case in our data), it has lower precision. The \emph{Best Match} policy performs well when there is a single match, but is not able to cope with arguments that have no matches or have multiple matches. The \emph{BM+Threshold} method combines the two and is useful when there are no matching key points or a single matching key point, but still have lower recall when there are multiple matching key points. The \emph{Dual Threshold} method improves the recall and therefore the F1 score for multiple matches while maintaining good performance for arguments with single or no matches.

Figure ~\ref{tab:prec-acc} shows Precision-Recall trade-off for the various policies, using the different possible thresholds, computed for one of the folds. For each policy, we specify the best F1 score, as well as the F1 score obtained for the selected threshold, which was optimized over the development set. The \emph{Threshold} policy allows to control recall, up to one (where the threshold is zero), at the price of low precision. The \emph{BM+Threshold} policy generates the highest precision, but 
low recall, since at most one candidate is selected. Note that when the threshold is zero, the \emph{BM+Threshold} policy is equivalent to the \emph{BM} policy. The \emph{Dual Threshold} policy offers the best trade-off, for mid-range precision and recall.
\section{Conclusion}
This work addressed the practical problem of summarizing a large collection of arguments on a given topic.
We proposed to represent such summaries as a set of key points scored according to their relative salience. Such summary aims to provide both textual and quantitative views of the argument data in a concise form. We demonstrated the feasibility and effectiveness of the proposed approach through extensive data annotation and analysis. We showed that a domain expert can quickly come up with a short list of pro and con key points per topic, that would capture the gist of crowd-contributed arguments, even without being exposed to the arguments themselves. We studied the problem of automatically matching arguments to key points, and developed the first large-scale dataset for this task, which we make publicly available.

Our experimental results demonstrate that the problem is far from trivial, and cannot be effectively solved using unsupervised methods based on word or sentence-level embedding. However, by using state of the art supervised learning methods for match scoring, together with an appropriate key point selection policy for match classification, we were able to achieve promising results on this task. 

The natural next step for this work is the challenging task of automatic key point generation. In addition, we plan to apply the methods presented in this work also to automatically-mined arguments. Finally, detecting the more implicit relations between the argument and the key point, as seen in our error analysis, is another intriguing direction for future work. 
\section*{Acknowledgments}
We would like to thank the anonymous reviewers for their helpful comments and suggestions.
\bibliography{keypoints}
\bibliographystyle{acl_natbib}
\end{document}